\documentclass[a4paper]{svproc}

\usepackage{url}

\usepackage{graphics} 
\usepackage{epsfig} 
\usepackage{hyperref}
\usepackage{mathptmx} 
\usepackage{times} 
\usepackage{amsmath} 
\usepackage{amssymb}  
\usepackage{todonotes}
\usepackage{booktabs}
\usepackage{multicol}
\usepackage[switch,columnwise]{lineno}
\usepackage{wrapfig} 
\usepackage{multirow}
\graphicspath{{image/}}
\usepackage[table]{xcolor}
\definecolor{lightgreen}{RGB}{200, 255, 200}  
\usepackage{todonotes}
\usepackage{caption}
\usepackage{soul}
\usepackage{svg}
\usepackage{algorithm2e}
\usepackage{float}
\usepackage{subcaption}

\usepackage{array}
\usepackage{tabularx}
\usepackage{wrapfig}
\usepackage{makecell}
\newcolumntype{Y}{>{\raggedright\arraybackslash}X} 
\usepackage{afterpage} 
\definecolor{mycolor}{HTML}{BBCCBA}
\usepackage{svg}

\usepackage{enumitem}

\definecolor{OrangeRed}{HTML}{FF4500}
\definecolor{DodgerBlue}{HTML}{1E90FF}
\definecolor{MediumSeaGreen}{HTML}{3CB371}
\definecolor{Gold}{HTML}{B8860B}
\newcommand{\qq}[1]{\(\text{CART}\)}

\usepackage{algpseudocode}

\title{\LARGE \bf
CART: Context-Aware Terrain Adaptation using Temporal Sequence Selection for Legged Robots
}

\usepackage{url}

\begin{document}
\mainmatter              
\title{CART: Context-Aware Terrain Adaptation using Temporal Sequence Selection for Legged Robots}
\titlerunning{Context-Aware Terrain Adaptation for Legged Robots}  
%
\author{
  Kartikeya Singh, Youngjin Kim, Yash Turkar, and Karthik Dantu\\
}

\authorrunning{Singh et al.} 

\institute{DRONES LAB, University at Buffalo, NY, USA,\\
\email{ksingh35@buffalo.edu} }

\maketitle

\thispagestyle{empty} 
\pagestyle{headings}  

\begin{abstract} 
Animals in nature combine multiple modalities, such as sight and feel, to perceive terrain and develop an understanding of how to walk on uneven terrain in an efficient manner. 
Similarly, legged robots need to develop their ability to stably walk on complex terrains by developing an understanding of the relationship between vision and proprioception. Most current terrain-adaptation methods remain susceptible to failure on complex off-road terrain because they do not explicitly model the context between exteroceptive terrain appearance and proprioceptive physical interaction. This experience-based learning often creates a Visual-Texture Paradox between what has been seen and how it actually feels. In this work, we introduce \qq{}, a high-level controller built on a context-aware terrain adaptation approach that integrates proprioception and exteroception from onboard sensing to achieve a robust understanding of terrain.  We evaluate our method on multiple terrains using the Unitree Go2 and ANYmal-C robot on the IsaacSim simulator and a Boston Dynamics SPOT robot for our real-world experiments. To evaluate whether the learned context improves locomotion behavior under the various paradox circumstances, we measure the robot's stability, traversal success, and task completion time in both simulation and real-world experiments. We compare CART against state-of-the-art locomotion
and terrain-adaptation baselines across diverse terrain conditions. CART improves the average success rate by \textbf{5\%} over the baselines in simulation, while improving context-conditioned locomotion behavior, including up to \textbf{41\%} lower
base oscillation in simulation and \textbf{22\%} in the real world, without increasing the time required to complete the locomotion tasks. 


\end{abstract}

\begin{figure*}
    \centering
    \vspace{-1cm}
    \includegraphics[width=\linewidth]{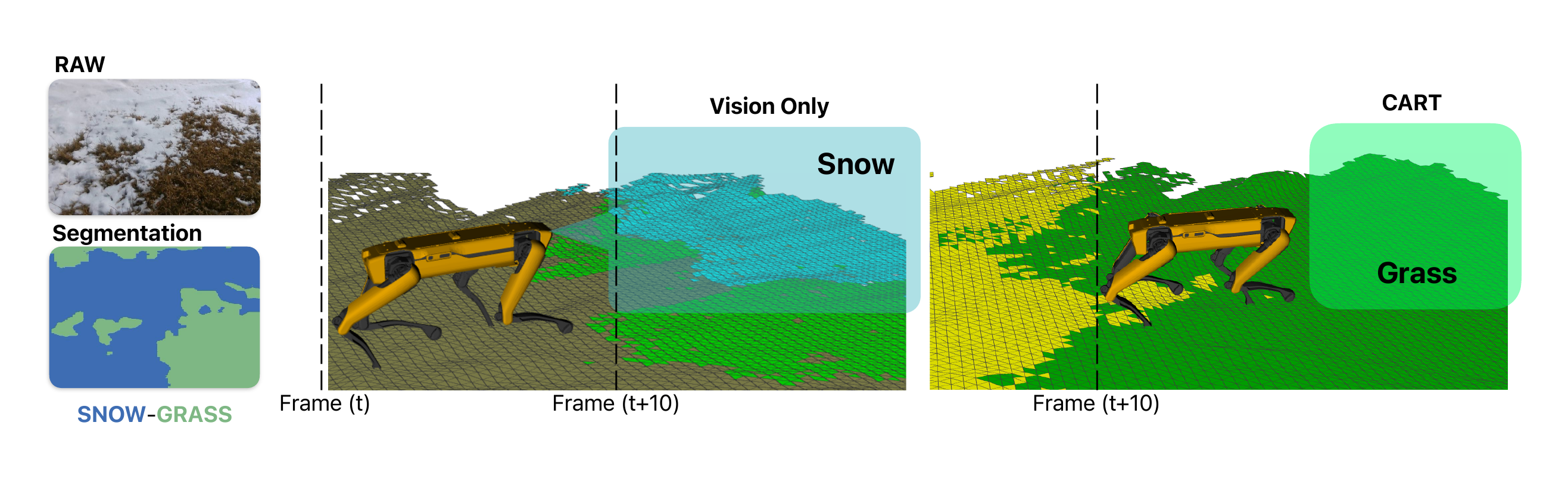}
    \caption{\textbf{Context switching in the real world using \qq{}:} Robot traverse through slushy snow overlayed on grass. On the left, we show the actual image projection from the robot's perspective that describes the raw image and snow-grass segments using \cite{viswanath2021offseg}. We show a \textit{Visual-Texture paradox} instance between vision only~\cite{ewen2022these} and \qq{}'s estimated category.}
    \vspace{-2cm}
    \label{context-snowgrass}
\end{figure*}


\section{INTRODUCTION}\label{intro}
Autonomous navigation of legged robots through diverse and complicated environments has gained attention over the past few years to perform critical tasks such as payload delivery~\cite{figliozzi2020autonomous}, search and rescue~\cite{bellicoso2018advances}, environmental inspection~\cite{kolvenbach2019haptic}, precision agriculture~\cite{naik2016precision}, and others. Legged robots such as Boston Dynamics' SPOT~\cite{bostondynamics2023}, Unitree Go1/Go2~\cite{unitree}, and Anybotics ANYmal~\cite{hutter2016anymal} provide basic legged locomotion, enabling them to walk on different terrains. The primary challenge in using them for off-road navigation is to incorporate a detailed understanding of the terrain and use this for high-level planning and control for safe and stable locomotion.

Several prior works use camera images to extract semantic information from off-road terrain~\cite{viswanath2021offseg}, \cite{guanganav}, \cite{zhong2022off}, \cite{yang2023wait}. These methods can identify terrain categories across outdoor environments, but image-based predictions can be sensitive to illumination, viewpoint, and appearance changes. Other approaches use LiDAR point clouds~\cite{viswanath2023off}, \cite{dabbiru2020lidar}, which are less affected by lighting and provide geometric terrain structure. Sensor-fusion methods further combine multiple exteroceptive modalities~\cite{kim2024ufo}, \cite{gao2021fine} to obtain richer terrain representations. However, exteroceptive terrain labels or geometry alone do not fully describe how a legged robot will physically interact with the surface. In particular, they may not capture traction, compliance, or contact-induced disturbances that affect locomotion behavior.

Sim-to-real approaches~\cite{agarwal2023legged}, \cite{fu2022coupling} have explored jointly using appearance and physical interaction cues for legged locomotion. These methods show that proprioceptive feedback is important for adapting to terrain properties such as traction, compliance, and support. Prior navigation methods have used exteroceptive sensing for terrain-aware planning in complex environments~\cite{sathyamoorthy2023using}, \cite{weerakoon2023adventr}, \cite{weerakoon2022graspe}, while recent terrain-adaptation methods combine exteroception with proprioceptive feedback for more stable locomotion~\cite{elnoor2024pronav}, \cite{weerakoon2024vapor}. However, experience-based policies can still struggle when visual appearance and physical interaction disagree, which we refer to as the \textit{Visual-Texture Paradox}. In such cases, a terrain may look like one class but produce contact behavior associated with another. Terrain adaptation, therefore, requires more than assigning a fixed terrain label from appearance. The robot must learn a context that links exteroceptive terrain cues with proprioceptive feedback from contact and motion. Figure~\ref{context-snowgrass} shows this mismatch in a snow-over-grass setting, where the robot visually observes snow but physically experiences grass-like support during traversal.

We present \qq{}, a context-aware terrain adaptation framework that learns this relationship and uses it to select high-level command sequences, including velocity scale and body height. Once the context is established, \qq{} selects commands that better match the current robot--terrain interaction, reducing foot slip and base oscillation/vibrational stability while maintaining traversal efficiency. Unlike single-modality approaches, \qq{} dynamically leverages exteroception and proprioception based on which modality is most informative for terrain assessment and locomotion adaptation.

The main contributions of this work are as follows:
\begin{itemize}[leftmargin=1em,topsep=0.1em,itemsep=0.05em]
  \item We introduce \qq{}, a context-aware terrain adaptation framework that combines
  exteroceptive terrain cues with proprioceptive feedback from the robot's physical
  interaction with the terrain.

  \item We propose a Temporal Sequence Selection mechanism that uses the learned
  context to select high-level locomotion commands, including velocity and body
  height, for stable traversal over complex terrains.

  \item We validate the learned context through downstream locomotion outcomes, including base stability,
  foot slip, traversal success, and task completion time. In simulation, \qq{} improves
  success rate by \textbf{5\%} and stability by up to \textbf{41\%}; in real-world
  experiments, it improves stability by up to \textbf{22\%} without increasing task
  completion time.
\end{itemize}

\section{Related Work}\label{sec:related_work}

\noindent{\textbf{Terrain Perception for Legged Navigation:}} Legged robots' navigation capability has been validated~\cite{wermelinger2016navigation}, \cite{agarwal2023legged} across complex, uneven, and varying terrain while ensuring safety and efficiency. With the hardware advancements in the industry, these robots have not been left untouched from performing some of the extreme navigation tasks like~\cite{wellhausen2023artplanner}, \cite{fu2022coupling}. Perpetual observation of terrain is a vital step in making the robust and collision-free navigation of a robot successful. Sensor-based perception using exteroception has been exploited over the years, which could result in a high-level but coarse characterization of a specific terrain. Leveraging sensor-based perception works like~\cite{guanganav}, and \cite{sathyamoorthy2024convoi} utilizes images to determine risk-aware path planning for a legged robot. However, these methods, without any additional signal, result in a lack of internal feedback and could make the robot's body unaware of its current state. To overcome these issues,~\cite{elnoor2024pronav} couples both exteroceptive and proprioceptive parameters of a robot to adapt between different gaits. Similarly, RL-based methods like~\cite{lee2020learning}, \cite{agarwal2023legged} generalize over different types of legged robots by adapting the various privileged information like simulated terrain profile, friction coefficients, and robot proprioception, but they are not well acquainted with the real-world terrain representations because of the sim-to-real distinction.

\noindent{\textbf{Base Motion Disturbances and Sensor Stability:}} Legged robots like Boston Dynamics' Spot and ANYMAL-C are designed to traverse complex, uneven terrain. However, these highly actuated robots are susceptible to creating intense, high-frequency motion jitteriness at the robot's base (pitch, roll, yaw). Tightly coupled sensors like camera/LiDAR inherit vibrations occurring on the base of the robot, leading to severe image blur or point cloud distortion~\cite{chen2024design}, inefficient localization and mapping~\cite{neuhaus2018mc2slam}\cite{taketomi2017visual} and descriptor matching \cite{gojcic2019perfect}. An entire domain of research in image and video stabilization is dedicated to addressing these issues. \cite{bazeille2017active} presents a stabilized vision system by using a fast pan and tilt unit (PTU) on the fully torque-controlled hydraulically actuated quadruped robot (HyQ) to compensate for the vibrations, impacts, or slippages on rough terrain. On the other hand,~\cite{wangy2025terrain} tackles the problem of SLAM degradation due to vibrations in the coupled sensors in rugged terrains by proposing a hybrid gimbal system with virtual suspension control and terrain-adaptive planning. Therefore, our work stabilizes the vibrations/oscillations on the robot's base to compensate for vibrations occurring in the coupled sensors, such as cameras/LiDAR mounted on the robot.

\noindent{\textbf{Learning-Based Locomotion Control:}}\label{vibration_stability_related_work} Model-based control approaches integrate an articulated dynamics robot model by leveraging physics and bio-inspired modeling techniques~\cite{blickhan1989spring},~\cite{ijspeert2001connectionist}. However, the constrained range of motion and non-adaptability in the robot's state bring suboptimal outcomes in robust-legged locomotion. The rise of deep reinforcement learning and the hardware evolution of legged robots has been complemented by simulation software like MuJoCo~\cite{todorov2012mujoco}, PyBullet~\cite{pybullet}, and Isaac Sim~\cite{isaacsim}. These simulations are capable of providing real-world-like environments that can be used to model different high-level controllers for a legged robot. A typical Markov Decision Process (MDP) can be obtained using the privileged sensory-based observation directly from a simulator. These observations include multimodal sensing and proprioceptive sensors (Joint positions, velocity, and effort) that could be used to formulate a reward function for policy training of legged robot locomotion, including Proximal policy optimization (PPO) algorithms~\cite{schulman2017proximal}, Soft actor-critic (SAC)~\cite{haarnoja2018soft}), and student-teacher (S\&T)~\cite{miki2022learning}. These handcrafted policies have shown improvements in learning stable locomotion on varying terrains. While SLR~\cite{chen2406slr} highlights the importance of proprioception for terrain understanding, exteroception remains important for anticipating terrain changes along the forward horizon. However, experience-based learning often fails to optimally estimate the true nature of the terrain due to the out-of-distribution terrain properties.

\begin{figure*}
    \centering
    \includegraphics[width=.9\linewidth,trim={20pt 10 20 10},clip]{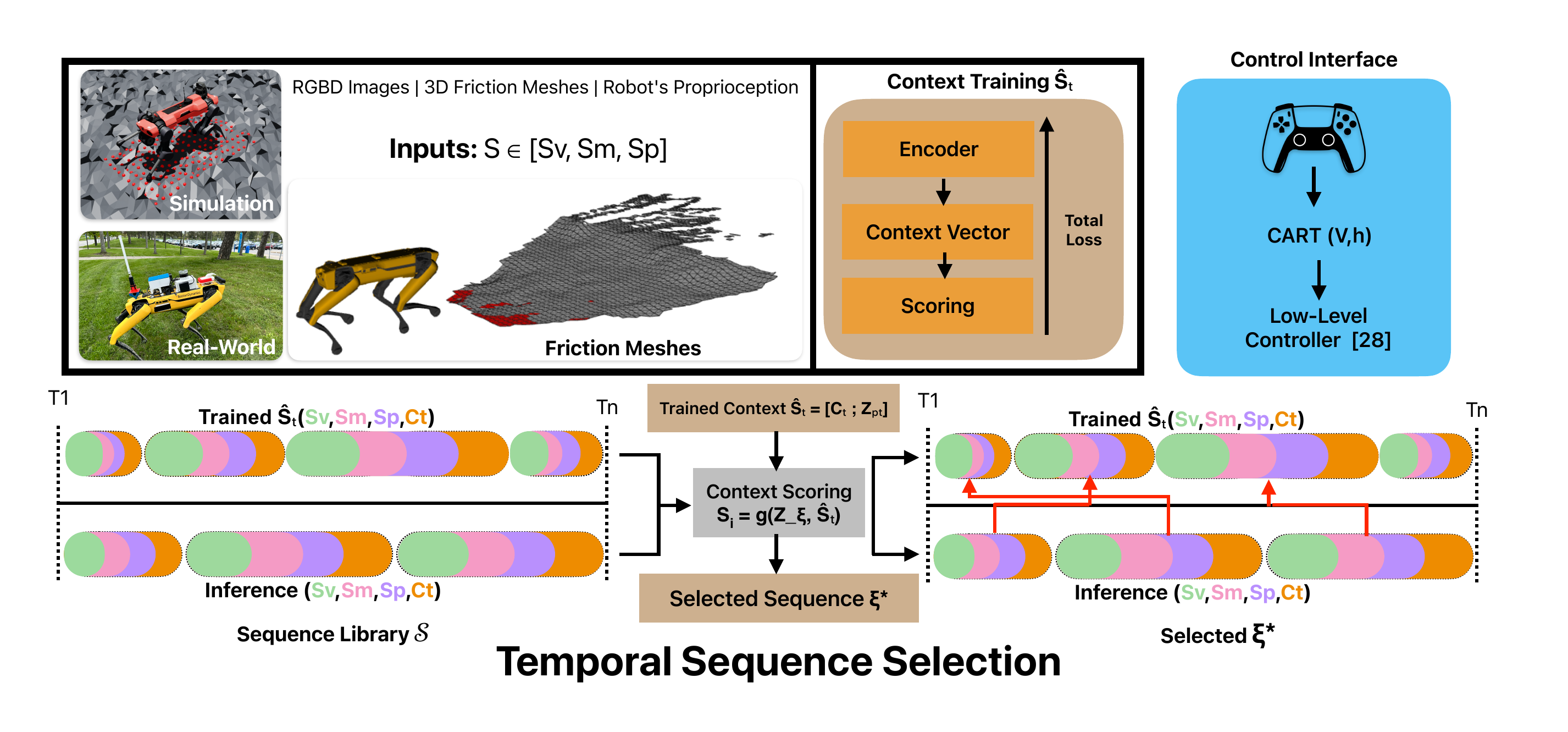}
    \vspace{-0.5cm}
    \caption{
Overview of CART. Multimodal observations from visual terrain cues $S_v$, terrain
geometry $S_m$, and proprioception $S_p$ are encoded into a context representation
$\hat{S}_t=[c_t;z_{p,t}]$. CART uses this context to adapt the high-level velocity scale
and body-height commands. During inference, TSS scores an offline library of short
command sequences $\xi_i=\{[\alpha,\Delta h]\}_{i:i+\ell}$ and selects the sequence
most consistent with the current robot--terrain interaction. The end-to-end framework is supervised using a rollout cost-based loss (\autoref{eq:rollout_cost}).
} 
    
    \label{overview}
    
\end{figure*}

\section{Method}\label{sec:method}

\noindent\textbf{Motivation: Why Context-Aware Terrain Adaptation?}
In this work, context refers to the relationship between what the robot observes about
the terrain and how the terrain physically affects the robot during contact. For legged
locomotion, terrain appearance alone is often incomplete or misleading. A surface may
look visually similar across regions while producing different traction, compliance, or
support, and exteroceptive sensing may also become unreliable due to occlusion, blur,
lighting, dust, or missing depth measurements. Similarly, proprioception alone can only
explain the terrain after the robot has already interacted with it. Therefore, robust
terrain adaptation requires learning a context that connects exteroceptive terrain cues
with proprioceptive interaction feedback.

Context-aware learning allows CART to move beyond assigning a fixed terrain label
from appearance. Instead, CART learns whether the robot's physical interaction agrees
or disagrees with the observed terrain cues, as in the snow-over-grass visual-texture
paradox, and whether useful proprioceptive structure remains when exteroception is
temporarily unavailable, as in the blind-phase setting. Once this context is learned,
CART uses it to adapt high-level command outputs, including velocity scale and
body-height offset. These context-conditioned commands allow the robot to reduce
undesired slip and base oscillation while maintaining traversal efficiency, leading to
more reliable locomotion behavior across complex terrain.

\noindent\textbf{Vibrational stability.} In this work, ``stability'' refers to \emph{vibrational stability} on the robot base: reducing oscillatory disturbances induced by foot--ground interaction that propagate to the base and can degrade onboard sensing (camera/depth) and perception-action coupling detailed in~\autoref{vibration_stability_related_work}. 

\noindent\textbf{Controller interface.}
CART operates hierarchically above an existing low-level locomotion controller. It does
not output joint-level actions; instead, it regulates high-level command parameters,
such as velocity scaling and body-height offset, around a given reference command. The
underlying controller then tracks the resulting base command using joint-level
actuation. Unless stated otherwise, all experiments in this work use the same pretrained low-level RL locomotion policy [28] to execute joint-level actions. On Spot, this mapping is handled by the built-in controller since the low-level control isn't publicly available for the Spot robot. This makes the CART
controller-agnostic as long as the robot exposes a base-command interface.

\subsection{Framework Overview}
CART combines multimodal context learning with temporal command selection for terrain-adaptive locomotion. First, a context representation is learned from visual, geometric, and proprioceptive observations (Section~\ref{sec:context_learning}). This context is then used to evaluate candidate velocity-scaling and body-height adaptations through a rollout-cost objective. Finally, the Temporal Sequence Selection (TSS) module (Section~\ref{sec:tss}) leverages the learned context to select command sequences that are executed by the underlying low-level locomotion controller.

\vspace{-0.5cm}

\subsection{Multimodal Context Embedding and Learning}\label{sec:context_learning}

\label{sec:context_fusion}
\vspace{-0.1cm}
\setlength{\parskip}{0pt}
At each timestep $t$, the robot observes $S_t = [S_{v,t},\, S_{m,t},\, S_{p,t}]$, where $S_{v,t}$ denotes RGB-D observations, $S_{m,t}$ denotes terrain geometry features, and $S_{p,t}$ denotes proprioceptive feedback. The purpose of the context module is to learn a compact representation that relates terrain appearance and
geometry to the robot's physical interaction with the terrain.

\subsubsection{Visual Encoding}
\vspace{-0.5cm}
\noindent We encode the visual observations $S_{v,t}=I_{RGBD,t}\in\mathbb{R}^{4\times360\times640}$
using a convolutional encoder
$z_{v,t}=f_v(S_{v,t})=f_v(I_{RGBD,t})\in\mathbb{R}^{256}$,
where $f_v(\cdot)$ consists of three convolutional layers with ReLU activations, adaptive average pooling, and a linear projection to 256 dimensions.

\subsubsection{Mesh Encoding}
\vspace{-0.5cm}
\noindent We encode the geometric observations $S_{m,t}=X_{mesh,t}\in\mathbb{R}^{128}$
using a two-layer MLP:
$z_{m,t}=f_m(S_{m,t})=f_m(X_{mesh,t})\in\mathbb{R}^{256}$.

\subsubsection{Proprioceptive Encoding}
\vspace{-0.5cm}
\noindent We encode proprioception via a bidirectional LSTM: $z_{p,t} = f_p(S_{p,t}) \in \mathbb{R}^{256}$.

\vspace{-0.5cm}

\subsubsection{Attention-Based Context Vector}

To explicitly model the relationship between terrain appearance/geometry and physical interaction feedback (thereby addressing the visual--texture paradox), we fuse exteroceptive latents using an attention mechanism:
$
c_t = \mathrm{Attn}(z_{v,t}, z_{m,t}) \in \mathbb{R}^{256}.
$
In our implementation, $\mathrm{Attn}(\cdot)$ is realized via multi-head attention applied to the pair $(z_{v,t}, z_{m,t})$, followed by aggregation. The final context-aware representation is formed by concatenating the fused exteroceptive context with the proprioceptive embedding:
$
\hat{S}_t = [c_t;\, z_{p,t}] \in \mathbb{R}^{512}.
$
This compact representation is used both for high-level command generation and for TSS context scoring. The learned context is trained to support high-level command selection. For each
terrain-context sample,
Candidate command profiles are evaluated using rollout costs computed from the robot's
measured behavior. For a candidate command
\begin{equation}
    u_t = [\alpha_t, \Delta h_t],
\end{equation}
where $\alpha_t$ is the velocity scale and $\Delta h_t$ is the body-height offset, the
resulting high-level command is
\begin{equation}
    v_t^{\mathrm{cmd}} = \alpha_t v_t^{\mathrm{ref}}, \qquad
    h_t^{\mathrm{cmd}} = h_0 + \Delta h_t .
\end{equation}

Each candidate command profile is assigned a rollout cost computed from the robot's
measured behavior as an end-to-end loss defined as:
\begin{equation}
    \mathcal{J}_t =
    w_{\Delta q}\ell_{\Delta q}(t)
    + w_{\mathrm{track}}\ell_{\mathrm{track}}(t)
    + w_{\mathrm{effort}}\ell_{\mathrm{effort}}(t),
    \label{eq:rollout_cost}
\end{equation}
where lower cost is better. The first term penalizes stance-foot displacement, the
second term preserves velocity tracking, and the third term discourages unnecessary
actuation effort. The individual terms are:
\begin{align}
    \ell_{\Delta q}(t) &=
    \sum_{\ell}
    \mathbf{1}_{\mathrm{stance}}(t,\ell)
    \left\|
    q^{\mathrm{foot}}_{t,\ell,xy}
    -
    q^{\mathrm{foot}}_{t-1,\ell,xy}
    \right\|_2,
    \label{eq:delta_q_cost} \\
    \ell_{\mathrm{track}}(t) &=
    \left\|
    v^{\mathrm{cmd}}_{t,xy} - v^{\mathrm{base}}_{t,xy}
    \right\|_2,
    \label{eq:tracking_cost} \\
    \ell_{\mathrm{effort}}(t) &=
    \|\tau_t\|_2^2 .
    \label{eq:effort_cost}
\end{align}

Here, $q^{\mathrm{foot}}_{t,\ell,xy}$ is the horizontal position (hereafter referred to as \textit{foot slip} for simplicity) of foot $\ell$ from the base of the robot at time
$t$, $\mathbf{1}_{\mathrm{stance}}(t,\ell)$ indicates whether the foot is in contact
with the ground, $v^{\mathrm{base}}_{t,xy}$ is the measured horizontal base velocity,
$v^{\mathrm{cmd}}_{t,xy}$ is the commanded horizontal velocity, and $\tau_t$ is the
joint torque vector. The foot slip term $\ell_{\Delta q}$ penalizes horizontal foot motion from the base of the robot
during stance, since a stance foot should ideally remain fixed relative to the ground.

\begin{figure}[t]
    \vspace{-0.2cm}
    \centering
    \includegraphics[
        width=\linewidth,
        trim={30 40 30 40},
        clip
    ]{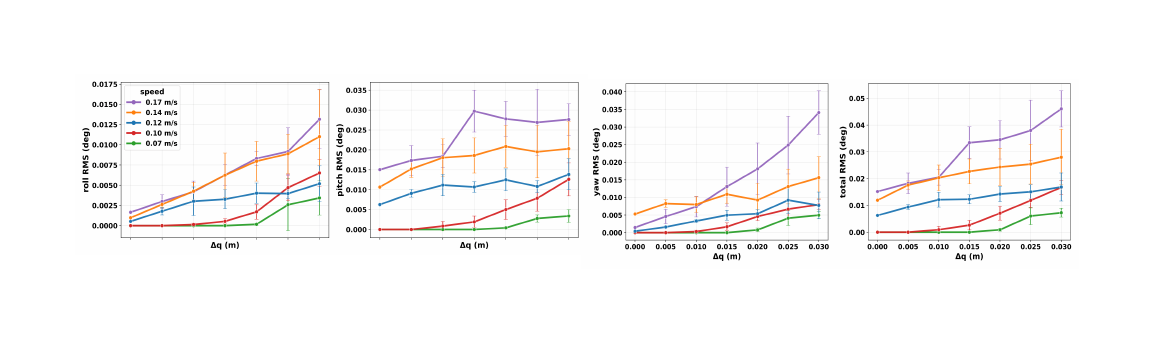}
    \vspace{-0.7cm}
    \caption{\textbf{Effect of lateral foot perturbation on base vibration:}
    Larger perturbation magnitude ($\Delta q$) generally increases base vibration across different walking speeds.}
    \label{fig:delta_qvsvibration}
    \vspace{-0.7cm}
\end{figure}

These rollout costs are used to rank candidate velocity-scale and body-height profiles.
CART is then trained to predict and select low-cost command profiles from the learned
context $\hat{S}_t$. Body vibration is not used as a direct loss term; instead, it is used
as an evaluation metric to verify whether reducing foot slip ($\Delta q$) and
selecting appropriate velocity commands also reduce oscillatory base motion. Figure~\ref{fig:delta_qvsvibration} motivates the relationship between foot-contact
disturbance and base vibration. For this study, we sweep commanded forward velocity
and inject lateral stance-foot offsets $\Delta q$ into the IK-generated foot targets,
then measure the resulting base roll, pitch, yaw, and angular-rate oscillations. The
results show that larger $\Delta q$ generally induces larger base vibration across
walking speeds, motivating the use of foot-contact disturbance as a signal for selecting
context-conditioned velocity and height commands.

\subsection{Temporal Sequence Selection (TSS)}
\label{sec:tss}

Temporal Sequence Selection (TSS) uses the learned context to choose how CART should
adapt the robot's high-level commands over a short future window. Rather than selecting
one command independently at every timestep, TSS selects a short command sequence
\(\xi_i\). Each sequence contains a velocity scale \(\alpha\) and a body-height offset
\(\Delta h\) at every step:
\begin{equation}
    \xi_i =
    \{[\alpha_i,\Delta h_i], \ldots,
    [\alpha_{i+\ell},\Delta h_{i+\ell}]\}.
\end{equation}
Here, \(\ell\) is the sequence length. The velocity scale \(\alpha\) adjusts the
reference velocity, while \(\Delta h\) raises or lowers the nominal walking height.

We build a library of candidate sequences, denoted by \(\mathcal{S}\), using different
velocity scales, height offsets, sequence lengths, and temporal overlaps. At runtime,
each candidate sequence \(\xi_i\) is embedded into a compact sequence feature
\(z_{\xi_i}\). TSS then scores this sequence feature together with the current context
\(\hat{S}_t\):
\begin{equation}
    s_i = g(z_{\xi_i}, \hat{S}_t),
\end{equation}
where \(s_i\) is the score for candidate sequence \(\xi_i\), and \(g(\cdot)\) is the
learned scoring network. The sequence with the highest score is selected:
\begin{equation}
    \xi_t^* = \arg\max_{\xi_i \in \mathcal{S}} s_i .
\end{equation}

CART applies only the first command from the selected sequence, then repeats the
selection process at the next timestep using the updated context (at the rate of 400ms per action frequency). If the first command
contains velocity scale \(\alpha_t\) and height offset \(\Delta h_t\), the high-level
commands sent to the low-level controller are:
\begin{equation}
    v_t^{\mathrm{cmd}} = \alpha_t v_t^{\mathrm{ref}}, \qquad
    h_t^{\mathrm{cmd}} = h_0 + \Delta h_t .
\end{equation}
Here, \(v_t^{\mathrm{ref}}\) is the reference velocity provided by the task, \(h_0\) is
the nominal body height, \(v_t^{\mathrm{cmd}}\) is the commanded velocity, and
\(h_t^{\mathrm{cmd}}\) is the commanded body height.

\section{Experiments and Results}\label{sec:evaluation}
\label{sec:experiments}

We evaluate CART in two stages. First, we test whether the learned context remains
useful under two complementary exteroceptive failure modes: Missing exteroception in the
blind-phase dropout setting and misleading exteroception
in the snow-over-grass visual-texture paradox. Second, we evaluate CART as an off-the-shelf high-level
terrain adaptation controller in simulation and real-world locomotion experiments.

\vspace{-0.5cm}

\subsection{Context Robustness Under Exteroceptive Failure}
\label{sec:context_failure_modes}
We first evaluate CART under two complementary exteroceptive failure modes. The first
is missing exteroception, where visual $S_v$ and geometric $S_m$ terrain cues are intentionally
masked during traversal. This controlled setting lets us evaluate whether the
proprioceptive component of the learned context $\hat{S}_t$ remains useful when exteroception is
unavailable. The second is misleading exteroception, where visual appearance is
available but does not match the robot's physical interaction with the terrain. Together,
these settings test whether CART can remain useful when exteroception is missing and
whether the learned context can resolve ambiguous visual terrain appearance using
proprioceptive interaction.

\textbf{Mode 1:} The first failure mode is the blind-phase setting in Isaacsim simulation~\autoref{fig:blind_phase_results} using Unitree Go2, where exteroceptive context ($S_v, S_m$) is
intentionally masked for fixed intervals during traversal. Here, the challenge is the missing exteroception that is required to maintain the context between the learned distribution of $S_v,S_m$ and $S_p$. This experiment evaluates whether
CART can continue adapting from proprioceptive context($S_p$ alone) when exteroceptive terrain cues are temporarily unavailable.

\textbf{Mode 2:}  The second failure mode is in the real-world using SPOT robot walking on the snow-over-grass visual-texture paradox shown in~\autoref{context-snowgrass}. In this setting, exteroception is available, but visual
appearance alone is misleading: snow-covered grass may be segmented as a single visual
class (Snow) even though the robot experiences different traction (Grass) and support as it moves
across snow, slush, and grass. This experiment evaluates whether CART can ground
ambiguous visual terrain cues using proprioceptive interaction and switch the context from snow to grass upon the stance.

\subsubsection{Blind-Phase Robustness Under Exteroceptive Dropout}
\label{sec:blind_phase_robustness}

To evaluate whether CART remains useful when exteroceptive sensing is temporarily
unavailable, we introduce a blind-phase experiment in which the robot traverses
multiple terrain segments while the exteroceptive context is masked over a fixed time
intervals. This setting is motivated by realistic deployment conditions such as temporary sensor failure. During these intervals, the robot cannot rely on visual or geometric
terrain cues, but it still receives proprioceptive feedback from its motion and contact
interaction with the ground.

In CART, the context representation is formed as
\begin{equation}
    \hat{S}_t = [c_t; z_{p,t}] \in \mathbb{R}^{512},
\end{equation}
where $c_t \in \mathbb{R}^{256}$ is the exteroceptive context obtained from visual
and geometric terrain cues, and $z_{p,t} \in \mathbb{R}^{256}$ is the proprioceptive
context. During blind intervals, we intentionally mask the exteroceptive component and use $\hat{S}^{\mathrm{blind}}_t = [0; z_{p,t}]$ for Temporal Sequence Selection (TSS). Therefore, the goal of this experiment is not
to show that CART reconstructs the missing exteroceptive context. Instead, it evaluates
whether the proprioceptive component $z_{p,t}$ remains informative enough for TSS to
continue selecting stabilizing velocity and height commands during exteroceptive
dropout.
We compare the IsaacGym baseline~\cite{rudin2022learning} against the same baseline
augmented with CART (Baseline+CART) over five terrain sequences. Each trial contains
alternating normal and blind intervals. During blind intervals, the 128-dimensional exteroceptive 
height-scan input of the baseline is masked to simulate exteroceptive dropout. For
Baseline+CART, the 256-dimensional exteroceptive component of the CART context is also masked (on top of 187 dimensions of the Baseline), and TSS
selects command sequences using the proprioceptive context:
$\hat{S}^{\mathrm{blind}}_t = [0; z_{p,t}]$.
We evaluate body vibration, foot slip, and traversal behavior.

\begin{figure}[t]
    \centering
    \includegraphics[width=\linewidth]{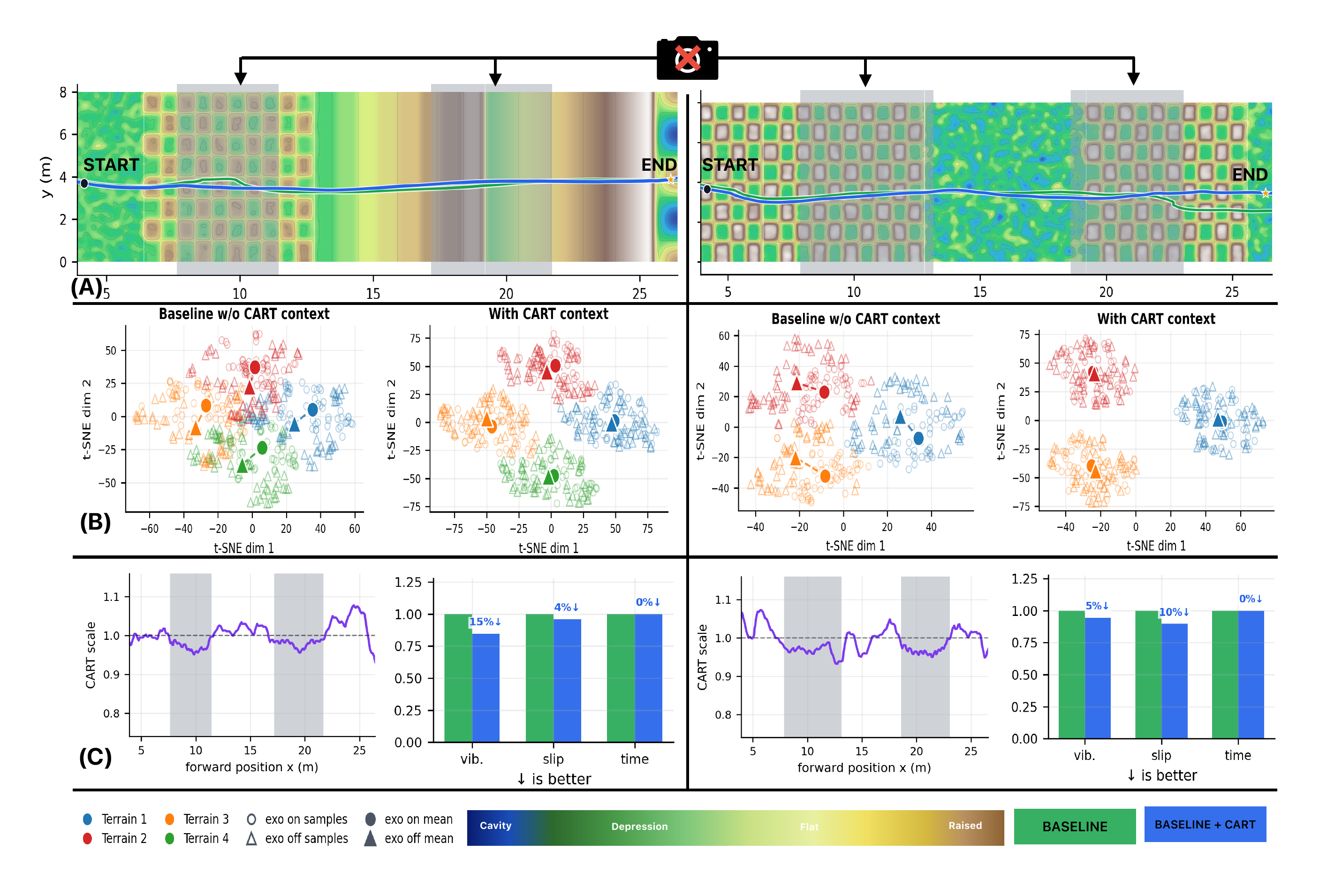}
    \vspace{-1cm}
    \caption{
\textbf{Blind-phase context robustness:} (A) Terrain height maps and robot trajectories for two
representative blind-phase runs. Gray regions indicate intervals where exteroception is disabled. (B) t-SNE
visualization of context embeddings with and without CART. Without CART, exteroceptive
dropout produces a larger shift between normal and blind samples. With CART, the
proprioceptive context remains grouped by terrain, and normal/blind samples stay close
within each terrain cluster. (C) CART-selected velocity scale during traversal and
normalized blind-phase metrics. Baseline~\cite{rudin2022learning}+CART reduces body vibration and foot slip while
maintaining comparable traversal time, indicating that CART improves blind-phase
locomotion behavior without relying on exteroception during masked intervals.
}
    \label{fig:blind_phase_results}
    \vspace{-0.5cm}
\end{figure}

~\autoref{fig:blind_phase_results} shows representative blind-phase runs with
randomly commanded linear and angular velocities. The height-map visualization confirms
that Baseline and Baseline+CART traverse the same terrain sequence under the same blind
intervals. The CART velocity-scale profile shows that TSS continues to adapt the
high-level command during blind intervals using proprioceptive context alone. This
adaptation primarily improves locomotion behavior rather than increasing speed:
Baseline+CART reduces oscillatory body motion in most blind intervals and often lowers
stance-foot slip compared with the baseline. Across the five terrain sequences,
Baseline+CART reduces body vibration in four out of five runs and reduces foot slip in
three out of five runs, while traversal time remains comparable. Thus, CART's
blind-phase benefit is best interpreted as improved locomotion behavior even under
exteroceptive dropout, rather than as a consistent increase in traversal speed.

The context embedding visualization in~\autoref{fig:blind_phase_results}(B)
provides additional qualitative evidence. For the baseline without CART, masking the
exteroceptive height-scan input produces a large representation shift between normal
and blind samples. With CART, blind and normal samples for the same terrain remain
closer in the learned context space, while samples from different terrains remain
separated. This indicates that CART does not collapse to a generic blind-state
representation. Instead, its proprioceptive context preserves terrain-relevant contact
information, supporting TSS-based command adaptation during blind traversal.

\subsubsection{Snow-over-Grass Visual-Texture Paradox}
\label{sec:snow_grass_paradox}

In the snow-over-grass experiment shown in~\autoref{context-snowgrass}, CART is deployed on the Spot robot in a real-world
terrain setting where visual appearance and physical interaction disagree. At the start
of the traversal, the robot observes a snow-covered surface, and the exteroceptive
context is therefore close to the snow cluster. However, as the robot walks over the
surface, the proprioceptive feedback reflects the underlying grass-like support and
traction. The learned CART context consequently shifts from the snow cluster toward the
grass cluster.

We analyze this behavior offline using the recorded context embeddings. No
exteroceptive masking is applied in this experiment. Instead, we use the same
context-embedding analysis used for the blind-phase study and compute representative
snow and grass context clusters from the recorded data. During traversal, each recorded
context embedding is assigned to the nearest cluster centroid. The assignment changes
from snow to grass as the robot moves over the snow-covered grass region, showing that
CART's context is not determined only by visual appearance. Rather, the context reflects
the agreement, or disagreement, between what the robot sees and what it physically
experiences through proprioception. Because real-world visual-texture paradoxes are difficult to reproduce systematically,
we next evaluate \qq{} as an off-the-shelf high-level controller in controlled simulation
and outdoor real-world experiments.

\subsection{CART as an Off-the-Shelf High-Level Controller}
\label{sec:cart_high_level_eval}

After evaluating context robustness under exteroceptive failure modes, we evaluate
\qq{} as an off-the-shelf high-level terrain adaptation controller. In this setting,
\qq{} regulates high-level locomotion commands around a given reference velocity,
including velocity scale and body-height offset, while the underlying robot controller
executes the resulting locomotion behavior. We evaluate this capability in simulation
using ANYmal-C in IsaacSim and in the real world using the Spot robot.

\subsubsection{Simulation Setup:}
\label{sec:sim_setup}

\begin{wrapfigure}{r}{0.52\columnwidth}
  
    \centering
    \includegraphics[width=0.50\columnwidth]{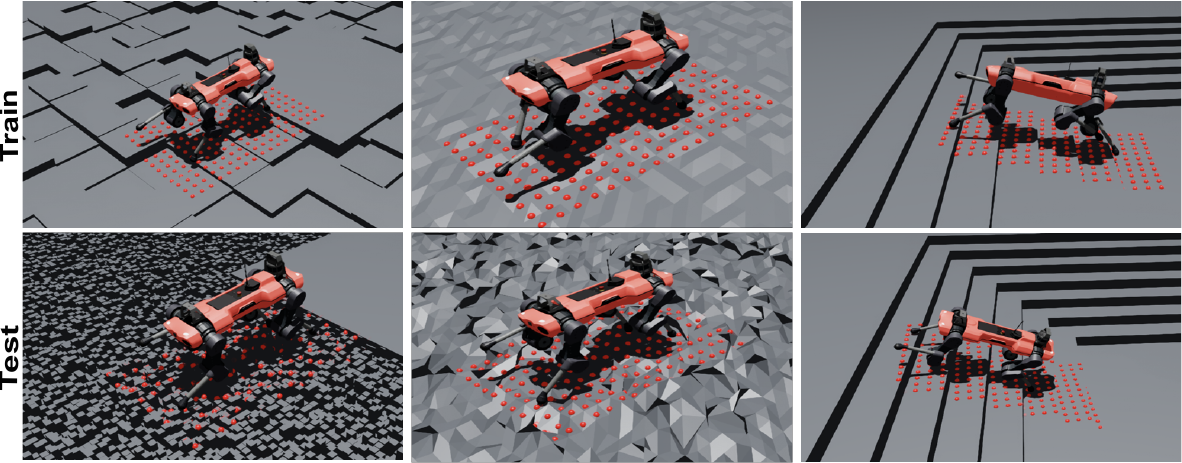}
    \caption{IsaacSim terrain configurations used for training and testing of all baselines along with \qq{} during our experiments. The top and bottom rows represent the difficulty of the same terrain type used for training and testing.}
    \label{fig:terrains}
    \vspace{-1cm}
\end{wrapfigure}

Our simulated experiments are conducted in IsaacSim~\cite{isaacsim} using the
ANYmal-C robot. We use IsaacLab's terrain-generation tools to construct deterministic
terrain configurations with varying difficulty. The simulated evaluation contains four
terrain categories, shown in~\autoref{fig:terrains}: box terrain, rough terrain, downward
slope, and upward slope. The box and rough terrains evaluate locomotion over flat but
irregular surfaces, while the slope terrains evaluate behavior over elevation changes.
All methods are trained on easier terrain settings and tested on harder terrain
configurations.

We compare CART against publicly available baselines grouped according to their sensing
modality. We use \textit{Proprioceptive Feedback (p)} to denote methods that rely only
on proprioceptive state, and \textit{Proprioceptive with Exteroceptive Feedback (p+e)}
to denote methods that use proprioception together with height scans, privileged
terrain information, or other exteroceptive terrain cues.

\noindent\textbf{Blind (p)}~\cite{cheng2024navila,stolle2024perceptive,sun2025learning}:
A proprioception-only locomotion policy trained using low-level robot state. This
policy does not use exteroceptive terrain information.

\noindent\textbf{Baseline (p+e)}~\cite{rudin2022learning}:
A Proximal Policy Optimization baseline that uses proprioceptive state together with
terrain features from an onboard height scanner as privileged terrain observations.

\noindent\textbf{S\&T (p+e)}~\cite{miki2022learning}:
A student-teacher locomotion policy that combines proprioception with privileged
exteroceptive terrain information during training and deploys a student policy for
terrain-aware locomotion.

\subsubsection{Real-World Setup}
\label{sec:real_world_setup}
\begin{wrapfigure}{r}{0.55\columnwidth}
    \vspace{-20pt}
    \centering
    \includegraphics[width=0.53\columnwidth]{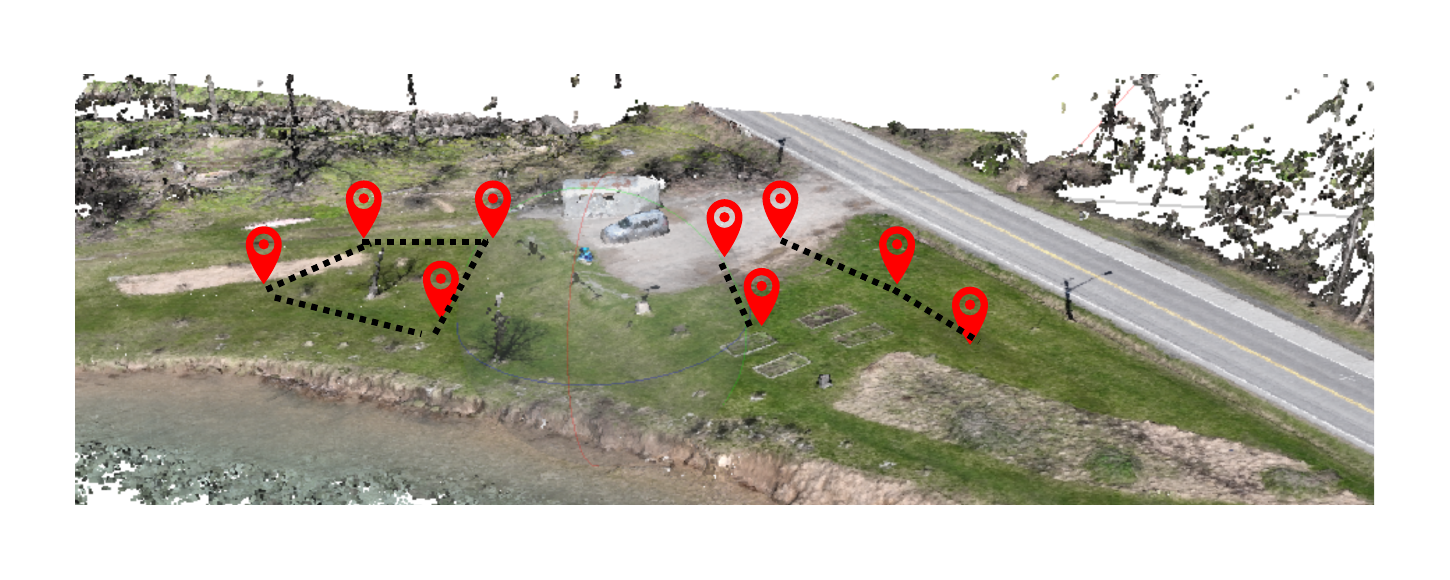}
    \vspace{-20pt}
    \caption{Real-world experiment scenario with multiple rugged terrains and elevations, including Grass, Mud, Concrete, Gravel, and Mulch. The experiments were conducted over various underlying terrains, such as Grass over Mud and Grass over Gravel. The red markers represent the waypoints, resulting in 7 runs per baseline.}
    \label{fig:lake}
    \vspace{-1cm}
\end{wrapfigure}

Our real-world experiments are conducted using a Boston Dynamics Spot robot on rugged
outdoor terrains with varying elevation, as shown in~\autoref{fig:lake}. The terrain
contains grass, mud, concrete, gravel, mulch, and mixed surfaces such as grass over mud
and grass over gravel. We follow the same point-to-point navigation schema used in
simulation.

\noindent\textbf{Spot Built-in Controller (p+e)}~\cite{bostondynamics2023}:
Spot's internal locomotion controller uses onboard sensing and feedback stabilization.
We compare against three built-in gait modes, \textit{TROT}, \textit{CRAWL}, and
\textit{AMBLE}, using identical commanded velocities.

\noindent\textbf{VAPOR (p+e)}~\cite{weerakoon2024vapor}:
A risk-aware navigation method that uses exteroceptive and proprioceptive feedback to
restrict the robot's velocity space for stable traversal in unstructured environments.
We use VAPOR's planner outputs and compare its stability and traversal behavior
against CART.

\begin{wrapfigure}{r}{0.50\columnwidth}
    \vspace{-1cm}
    \centering
    \includegraphics[width=0.48\columnwidth]{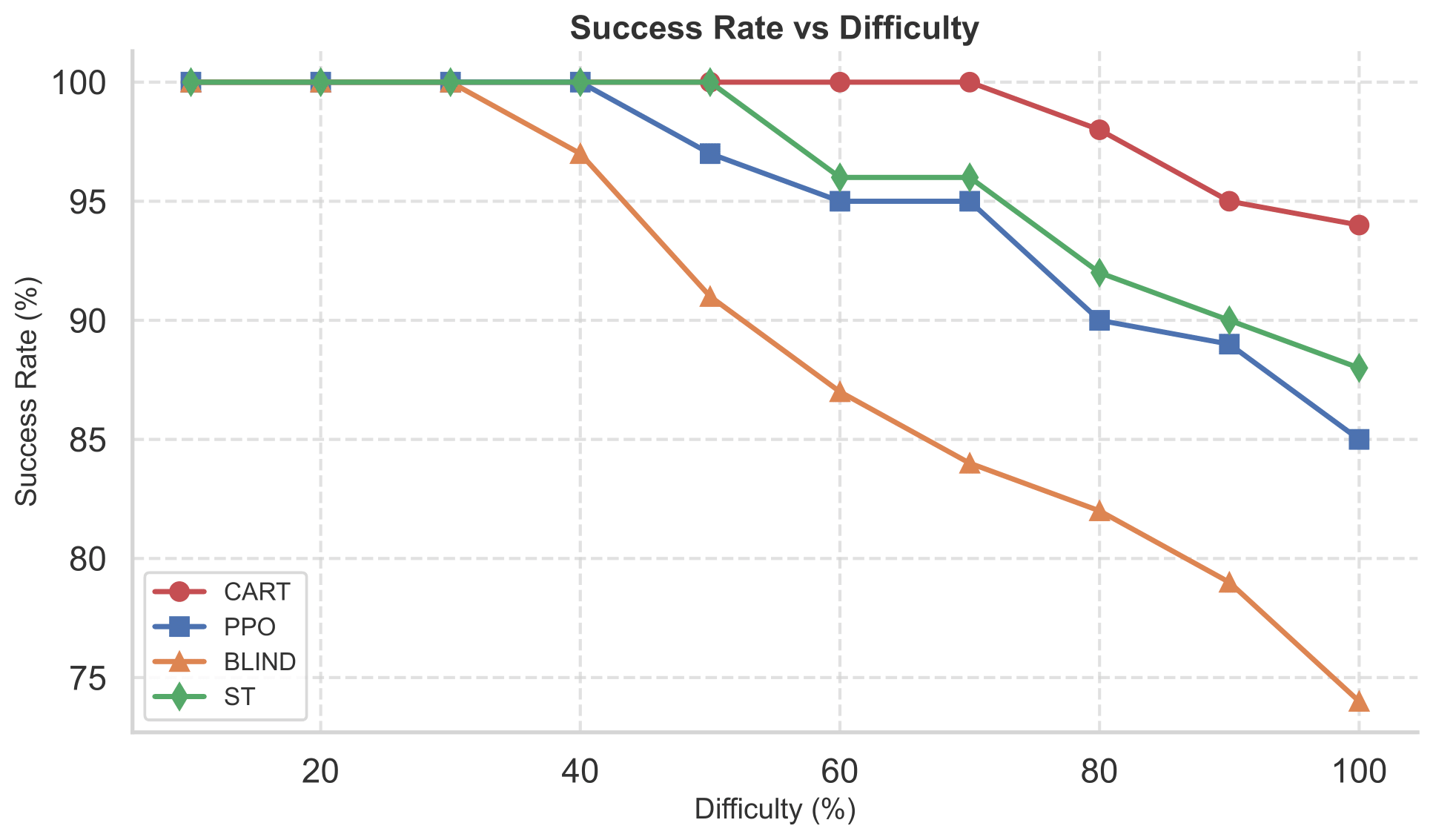}
    \vspace{-10pt}
    \caption{\textbf{Success rates:} 
    Success rates during simulation-based point-to-point traversal experiments across terrains of increasing difficulty. Results are reported over 100 independent trials per terrain.}
    \label{fig:success}
    \vspace{-20pt}
\end{wrapfigure}

\noindent\textbf{Data Collection and Setup:} Data for real-world training was collected across different terrains like grass, concrete, gravel, pebble sidewalks, etc. Whereas, the IsaacSim model was trained using terrain configurations shown in~\autoref{fig:terrains}. During the data collection for both simulation and real-world, we recorded High-volume data from the Spot's and ANYmal's joints (position and effort), hips (position and velocity), feet (slips), robot’s command velocity, and height. We used a Zed2i camera mounted on top of Spot's robot to collect RGBD images and a Microstrain IMU for the body oscillation evaluation. The data collected from the real world exhibited distinct visual characteristics like grass, gravel, mud, etc. However, the simulation terrain lacked this property and relied only on depth and proprioception. Therefore, \qq{} scores the selected sequences using proprioceptive and depth context only. This property of \qq{} couples the contexts from other terrains that the robot might acquire while walking.

\subsection{Evaluation Metrics and Results}

For both simulation and real-world experiments, we evaluate whether the learned context
improves downstream locomotion behavior. We measure base-motion disturbances,
traversal success in simulation, distance traveled in the real world, and task
completion time. These metrics test whether context-conditioned command selection
improves locomotion across platforms and terrain conditions.

\definecolor{lightgreen}{rgb}{0.88,1,0.88}
\begin{wraptable}{r}{0.42\columnwidth}
\vspace{-28pt}
\centering
\scriptsize
\setlength{\tabcolsep}{3pt}
\renewcommand{\arraystretch}{0.9}

\caption{\textbf{Performance across methods. Simulation results are averaged over 10 successful trials per terrain, while real-world results are averaged over 7 traversals per method using the experimental layout shown in~\autoref{fig:lake}}}
\vspace{-6pt}

\label{tab:distance-time}

\begin{tabular}{lc|lc}
\toprule
\multicolumn{2}{c|}{\textbf{Simulation}} &
\multicolumn{2}{c}{\textbf{Real World}} \\
Method & Time$\downarrow$ & Method & Dist.$\uparrow$ \\
\midrule
PPO   & 35.15 & TROT  & 4.72 \\
Blind & 49.45 & CRAWL & 2.49 \\
ST    & 31.91 & AMBLE & 4.44 \\
      &       & VAPOR & 4.66 \\
CART  & \cellcolor{lightgreen}\textbf{29.28}
      & CART & \cellcolor{lightgreen}\textbf{5.44} \\
\bottomrule
\end{tabular}

\vspace{-1cm}
\end{wraptable}

\noindent\textbf{Success Rates:} The success rates of the ANYmal robot while traversing through unseen and complex terrains in IsaacSim. The corresponding simulation results with 100 trials per terrain are illustrated in~\autoref{fig:success}. The locomotive tasks were defined to make the robot follow an intended path from one point to another on each given terrain. Since the source and goal were pre-defined, we evaluate the time taken by the robot to complete tasks in~\autoref{tab:distance-time}. We report the mean of
10 successful experiments, because the blind policy performed very poorly.

\begin{table*}[t]
\centering
\caption{\textbf{Simulation and Real-World Results:} Mean $\pm$ variance of body angles (rad) and angular rates (rad/s) across flat and elevated terrains. Simulation results are averaged over 10 independent trials per terrain, while real-world results are averaged over 7 traversals per method using the experimental layout shown in~\autoref{fig:lake}.}
\label{tab:combined-results}

\scriptsize
\setlength{\tabcolsep}{1.6pt}
\renewcommand{\arraystretch}{0.80}

\begin{tabular}{@{}llcccccc@{}}
\toprule
& & \multicolumn{2}{c}{\textbf{Roll}} & \multicolumn{2}{c}{\textbf{Pitch}} & \multicolumn{2}{c}{\textbf{Yaw}} \\
\cmidrule(lr){3-4}\cmidrule(lr){5-6}\cmidrule(lr){7-8}
\textbf{Scenario} & \textbf{Method}
  & \textbf{Angle} & \textbf{Rate}
  & \textbf{Angle} & \textbf{Rate}
  & \textbf{Angle} & \textbf{Rate} \\
\midrule

\multicolumn{8}{l}{\textit{Simulation}} \\
\midrule

\multirow{4}{*}{\makecell[l]{Flat}}
& PPO
  & $0.12{\pm}0.02$ & $0.56{\pm}0.59$
  & $0.06{\pm}0.00$ & $0.39{\pm}0.25$
  & $0.06{\pm}0.01$ & $0.15{\pm}0.04$ \\
& BLIND
  & $0.05{\pm}0.00$ & $0.59{\pm}0.57$
  & $0.15{\pm}0.04$ & $0.39{\pm}0.24$
  & $0.15{\pm}0.04$ & $0.18{\pm}0.08$ \\
& ST
  & $0.09{\pm}0.01$ & $0.66{\pm}0.78$
  & $0.06{\pm}0.01$ & $0.44{\pm}0.31$
  & $0.08{\pm}0.01$ & $0.14{\pm}0.04$ \\
& CART
  & \cellcolor{green!20}$\mathbf{0.03{\pm}0.00}$ & \cellcolor{green!20}$\mathbf{0.40{\pm}0.28}$
  & \cellcolor{green!20}$\mathbf{0.04{\pm}0.00}$ & \cellcolor{green!20}$\mathbf{0.29{\pm}0.14}$
  & \cellcolor{green!20}$\mathbf{0.02{\pm}0.00}$ & \cellcolor{green!20}$\mathbf{0.09{\pm}0.01}$ \\
\cmidrule(lr){1-8}

\multirow{4}{*}{\makecell[l]{Elev.}}
& PPO
  & $0.07{\pm}0.01$ & $0.61{\pm}0.70$
  & $0.23{\pm}0.07$ & $0.36{\pm}0.24$
  & $0.05{\pm}0.00$ & $0.12{\pm}0.03$ \\
& BLIND
  & $0.53{\pm}0.95$ & $0.99{\pm}1.74$
  & $0.23{\pm}0.07$ & $0.61{\pm}0.63$
  & $0.28{\pm}0.14$ & $0.36{\pm}0.20$ \\
& ST
  & $0.08{\pm}0.01$ & $0.67{\pm}0.76$
  & $0.18{\pm}0.05$ & $0.45{\pm}0.33$
  & $0.07{\pm}0.01$ & $0.14{\pm}0.03$ \\
& CART
  & \cellcolor{green!20}$\mathbf{0.05{\pm}0.00}$ & \cellcolor{green!20}$\mathbf{0.45{\pm}0.35}$
  & \cellcolor{green!20}$\mathbf{0.16{\pm}0.04}$ & \cellcolor{green!20}$\mathbf{0.36{\pm}0.19}$
  & \cellcolor{green!20}$\mathbf{0.06{\pm}0.00}$ & \cellcolor{green!20}$\mathbf{0.10{\pm}0.02}$ \\
\midrule

\multicolumn{8}{l}{\textit{Real World}} \\
\midrule

\multirow{5}{*}{\makecell[l]{Flat}}
& TROT
  & $0.03{\pm}0.00$ & $0.45{\pm}0.38$
  & $0.03{\pm}0.00$ & $0.15{\pm}0.04$
  & $0.10{\pm}0.03$ & $0.07{\pm}0.01$ \\
& CRAWL
  & $0.02{\pm}0.00$ & $0.30{\pm}0.10$
  & $0.03{\pm}0.00$ & $0.17{\pm}0.05$
  & $0.09{\pm}0.02$ & $0.05{\pm}0.01$ \\
& AMBLE
  & $0.02{\pm}0.00$ & $0.33{\pm}0.21$
  & $0.03{\pm}0.00$ & $0.14{\pm}0.04$
  & $0.09{\pm}0.04$ & $0.06{\pm}0.01$ \\
& VAPOR
  & $0.02{\pm}0.00$ & $0.36{\pm}0.25$
  & $0.03{\pm}0.00$ & $0.15{\pm}0.04$
  & $0.10{\pm}0.04$ & $0.06{\pm}0.01$ \\
& CART
  & \cellcolor{green!20}$\mathbf{0.01{\pm}0.00}$ & \cellcolor{green!20}$\mathbf{0.29{\pm}0.19}$
  & \cellcolor{green!20}$\mathbf{0.03{\pm}0.00}$ & \cellcolor{green!20}$\mathbf{0.13{\pm}0.04}$
  & \cellcolor{green!20}$\mathbf{0.08{\pm}0.02}$ & \cellcolor{green!20}$\mathbf{0.05{\pm}0.01}$ \\
\cmidrule(lr){1-8}

\multirow{5}{*}{\makecell[l]{Elev.}}
& TROT
  & $0.00{\pm}0.00$ & $0.12{\pm}0.03$
  & $0.10{\pm}0.02$ & $0.17{\pm}0.06$
  & $0.00{\pm}0.00$ & $0.04{\pm}0.00$ \\
& CRAWL
  & $0.02{\pm}0.00$ & $0.14{\pm}0.04$
  & $0.14{\pm}0.03$ & $0.20{\pm}0.07$
  & $0.00{\pm}0.01$ & $0.06{\pm}0.01$ \\
& AMBLE
  & $0.01{\pm}0.01$ & $0.14{\pm}0.04$
  & $0.11{\pm}0.03$ & $0.15{\pm}0.07$
  & $0.01{\pm}0.00$ & $0.05{\pm}0.01$ \\
& VAPOR
  & $0.01{\pm}0.01$ & \cellcolor{green!20}$\mathbf{0.12{\pm}0.05}$
  & $0.11{\pm}0.02$ & $0.18{\pm}0.06$
  & $0.00{\pm}0.00$ & $0.08{\pm}0.01$ \\
& CART
  & \cellcolor{green!20}$\mathbf{0.00{\pm}0.00}$ & $0.14{\pm}0.03$
  & \cellcolor{green!20}$\mathbf{0.10{\pm}0.01}$ & \cellcolor{green!20}$\mathbf{0.15{\pm}0.05}$
  & \cellcolor{green!20}$\mathbf{0.00{\pm}0.00}$ & \cellcolor{green!20}$\mathbf{0.04{\pm}0.00}$ \\
\bottomrule
\end{tabular}

\end{table*}

\noindent\textbf{Experimental Performance Analysis:} In order to account for the efficiency of the \qq{}, we analyze various parameters by measuring task completion time and total distance traveled. The simulated test runs were performed using predefined paths that the robot was intended to follow. However, in the real world, the planner was missing. Therefore, we calculate the total distance traveled by the robot on a particular terrain of long horizon in a given time. The corresponding evaluation metrics are tabulated in~\autoref{tab:combined-results}. Our experiments demonstrate better performance in both simulation and real-world implementation across all the compared baselines.

\noindent\textbf{Testing context-conditioned locomotion behavior:}
For both simulation and real-world experiments, we evaluate whether the learned context
improves downstream locomotion behavior. We measure base-motion disturbances,
traversal success in simulation, distance traveled in the real world, and task completion
time. These metrics test whether context-conditioned command selection improves
locomotion across platforms and terrain conditions. As motivated by~\autoref{fig:delta_qvsvibration}, we use base-motion disturbance as
one measure of locomotion quality. In the real-world experiments, we mount an
additional IMU near the center of the robot body to measure roll, pitch, yaw, and their
rates of change. In simulation, the same quantities are extracted from the robot state.
These signals capture orientation and angular-rate perturbations around the robot's
nominal body motion. The corresponding measurements are reported in
\autoref{tab:combined-results} for both simulation and real-world experiments. We compute percentage improvement as $\text{Avg. Improvement}(\%) = \frac{1}{M}\sum_{i=1}^{M}\left(\frac{\bar{X}_i-\bar{X}_{\text{CART}}}{\bar{X}_i}\right)\times 100$, where $M$ is the number of baselines, $\bar{X}_i$ is the mean metric value of the $i$-th baseline, and $\bar{X}_{\text{CART}}$ is the mean metric value of CART.

\section{Conclusion and Future Work}
\label{sec:conclusion}

We presented \qq{}, a context-aware terrain adaptation framework for robust quadruped locomotion over irregular and complex terrains. CART combines exteroceptive terrain cues with proprioceptive interaction feedback to adapt high-level locomotion commands, including desired velocity and body height, enabling stable traversal under challenging terrain conditions and visual ambiguities such as the visual-texture paradox. CART uses Temporal Sequence Selection (TSS) to select context-consistent
command sequences that reduce foot slip, oscillatory base motion, and
terrain-induced disturbances. Experimental results in both simulation and real-world environments demonstrate improved locomotion performance across diverse terrains. From our experiments, \qq{} improves average success rate by \textbf{5\%} and locomotion stability by \textbf{41}\% in simulation, while achieving \textbf{22}\% improvement in real-world stability on unseen outdoor terrains. The blind-phase experiments further
show that CART remains useful under temporary exteroceptive dropout. Even when visual
and geometric observations are unavailable, the proprioceptive context remains
informative enough for TSS to continue selecting stabilizing velocity and height
commands, reducing slip and base vibration during traversal.

Future work will extend CART beyond velocity and height adaptation toward richer
high-level and low-level control objectives, including adaptive gait selection,
contact-aware motion planning, and energy-efficient locomotion in large-scale outdoor
environments.
\bibliographystyle{splncs04}

\bibliography{references}

@article{figliozzi2020autonomous,
  title={Autonomous delivery robots and their potential impacts on urban freight energy consumption and emissions},
  author={Figliozzi, Miguel and Jennings, Dylan},
  journal={Transportation research procedia},
  volume={46},
  pages={21--28},
  year={2020},
  publisher={Elsevier}
}

@inproceedings{naik2016precision,
  title={Precision agriculture robot for seeding function},
  author={Naik, Neha S and Shete, Virendra V and Danve, Shruti R},
  booktitle={2016 international conference on inventive computation technologies (ICICT)},
  volume={2},
  pages={1--3},
  year={2016},
  organization={IEEE}
}

@article{bellicoso2018advances,
  title={Advances in real-world applications for legged robots},
  author={Bellicoso, C Dario and Bjelonic, Marko and Wellhausen, Lorenz and Holtmann, Kai and G{\"u}nther, Fabian and Tranzatto, Marco and Fankhauser, Peter and Hutter, Marco},
  journal={Journal of Field Robotics},
  volume={35},
  number={8},
  pages={1311--1326},
  year={2018},
  publisher={Wiley Online Library}
}

@article{kolvenbach2019haptic,
  title={Haptic inspection of planetary soils with legged robots},
  author={Kolvenbach, Hendrik and B{\"a}rtschi, Christian and Wellhausen, Lorenz and Grandia, Ruben and Hutter, Marco},
  journal={IEEE Robotics and Automation Letters},
  volume={4},
  number={2},
  pages={1626--1632},
  year={2019},
  publisher={IEEE}
}

@inproceedings{viswanath2021offseg,
  title={Offseg: A semantic segmentation framework for off-road driving},
  author={Viswanath, Kasi and Singh, Kartikeya and Jiang, Peng and Sujit, PB and Saripalli, Srikanth},
  booktitle={2021 IEEE 17th international conference on automation science and engineering (CASE)},
  pages={354--359},
  year={2021},
  organization={IEEE}
}

@article{guanganav,
  title={GANav: Group-wise Attention for Classifying Navigable Regions in Unstructured Outdoor Environments},
  author={Guan, Tianrui and Kothandaraman, Divya and Chandra, Rohan and Sathyamoorthy, Adarsh Jagan and Manocha, Dinesh}
}

@inproceedings{yang2023wait,
  title={Wait, that feels familiar: Learning to extrapolate human preferences for preference-aligned path planning},
  author={Yang, Elvin and Karnan, Haresh and Warnell, Garrett and Stone, Peter and Biswas, Joydeep},
  booktitle={ICRA2023 Workshop on Pretraining for Robotics (PT4R)},
  year={2023}
}

@article{kim2024ufo,
  title={UFO: Uncertainty-aware LiDAR-image Fusion for Off-road Semantic Terrain Map Estimation},
  author={Kim, Ohn and Seo, Junwon and Ahn, Seongyong and Kim, Chong Hui},
  journal={arXiv preprint arXiv:2403.02642},
  year={2024}
}

@inproceedings{gao2021fine,
  title={Fine-grained off-road semantic segmentation and mapping via contrastive learning},
  author={Gao, Biao and Hu, Shaochi and Zhao, Xijun and Zhao, Huijing},
  booktitle={2021 IEEE/RSJ International Conference on Intelligent Robots and Systems (IROS)},
  pages={5950--5957},
  year={2021},
  organization={IEEE}
}

@article{sathyamoorthy2023using,
  title={Using Lidar Intensity for Robot Navigation},
  author={Sathyamoorthy, Adarsh Jagan and Weerakoon, Kasun and Elnoor, Mohamed and Manocha, Dinesh},
  journal={arXiv preprint arXiv:2309.07014},
  year={2023}
}

@inproceedings{weerakoon2023adventr,
  title={Adventr: Autonomous robot navigation in complex outdoor environments},
  author={Weerakoon, Kasun and Sathyamoorthy, Adarsh Jagan and Elnoor, Mohamed and Manocha, Dinesh},
  booktitle={International Symposium on Experimental Robotics},
  pages={219--228},
  year={2023},
  organization={Springer}
}

@article{weerakoon2022graspe,
  title={Graspe: Graph based multimodal fusion for robot navigation in unstructured outdoor environments},
  author={Weerakoon, Kasun and Sathyamoorthy, Adarsh Jagan and Liang, Jing and Guan, Tianrui and Patel, Utsav and Manocha, Dinesh},
  journal={arXiv preprint arXiv:2209.05722},
  year={2022}
}

@inproceedings{agarwal2023legged,
  title={Legged locomotion in challenging terrains using egocentric vision},
  author={Agarwal, Ananye and Kumar, Ashish and Malik, Jitendra and Pathak, Deepak},
  booktitle={Conference on robot learning},
  pages={403--415},
  year={2023},
  organization={PMLR}
}

@inproceedings{fu2022coupling,
  title={Coupling vision and proprioception for navigation of legged robots},
  author={Fu, Zipeng and Kumar, Ashish and Agarwal, Ananye and Qi, Haozhi and Malik, Jitendra and Pathak, Deepak},
  booktitle={Proceedings of the IEEE/CVF Conference on Computer Vision and Pattern Recognition},
  pages={17273--17283},
  year={2022}
}

@article{elnoor2024pronav,
  title={Pronav: Proprioceptive traversability estimation for legged robot navigation in outdoor environments},
  author={Elnoor, Mohamed and Sathyamoorthy, Adarsh Jagan and Weerakoon, Kasun and Manocha, Dinesh},
  journal={IEEE Robotics and Automation Letters},
  year={2024},
  publisher={IEEE}
}

@inproceedings{wermelinger2016navigation,
  title={Navigation planning for legged robots in challenging terrain},
  author={Wermelinger, Martin and Fankhauser, P{\'e}ter and Diethelm, Remo and Kr{\"u}si, Philipp and Siegwart, Roland and Hutter, Marco},
  booktitle={2016 IEEE/RSJ International Conference on Intelligent Robots and Systems (IROS)},
  pages={1184--1189},
  year={2016},
  organization={IEEE}
}

@article{wellhausen2023artplanner,
  title={Artplanner: Robust legged robot navigation in the field},
  author={Wellhausen, Lorenz and Hutter, Marco},
  journal={arXiv preprint arXiv:2303.01420},
  year={2023}
}

@article{lee2020learning,
  title={Learning quadrupedal locomotion over challenging terrain},
  author={Lee, Joonho and Hwangbo, Jemin and Wellhausen, Lorenz and Koltun, Vladlen and Hutter, Marco},
  journal={Science robotics},
  volume={5},
  number={47},
  pages={eabc5986},
  year={2020},
  publisher={American Association for the Advancement of Science}
}

@inproceedings{weerakoon2024vapor,
  title={VAPOR: Legged Robot Navigation in Unstructured Outdoor Environments using Offline Reinforcement Learning},
  author={Weerakoon, Kasun and Sathyamoorthy, Adarsh Jagan and Elnoor, Mohamed and Manocha, Dinesh},
  booktitle={2024 IEEE International Conference on Robotics and Automation (ICRA)},
  pages={10344--10350},
  year={2024},
  organization={IEEE}
}

@article{sathyamoorthy2024convoi,
  title={CoNVOI: Context-aware Navigation using Vision Language Models in Outdoor and Indoor Environments},
  author={Sathyamoorthy, Adarsh Jagan and Weerakoon, Kasun and Elnoor, Mohamed and Zore, Anuj and Ichter, Brian and Xia, Fei and Tan, Jie and Yu, Wenhao and Manocha, Dinesh},
  journal={arXiv preprint arXiv:2403.15637},
  year={2024}
}

@inproceedings{viswanath2023off,
  title={Off-road lidar intensity based semantic segmentation},
  author={Viswanath, Kasi and Jiang, Peng and Sujit, PB and Saripalli, Srikanth},
  booktitle={International Symposium on Experimental Robotics},
  pages={608--617},
  year={2023},
  organization={Springer}
}

@article{dabbiru2020lidar,
  title={Lidar data segmentation in off-road environment using convolutional neural networks (cnn)},
  author={Dabbiru, Lalitha and Goodin, Chris and Scherrer, Nicklaus and Carruth, Daniel},
  journal={SAE International Journal of Advances and Current Practices in Mobility},
  volume={2},
  number={2020-01-0696},
  pages={3288--3292},
  year={2020}
}

@article{zhong2022off,
  title={Off-road drivable area detection: A learning-based approach exploiting lidar reflection texture information},
  author={Zhong, Chuanchuan and Li, Bowen and Wu, Tao},
  journal={Remote Sensing},
  volume={15},
  number={1},
  pages={27},
  year={2022},
  publisher={MDPI}
}

@inproceedings{hutter2016anymal,
  title={Anymal-a highly mobile and dynamic quadrupedal robot},
  author={Hutter, Marco and Gehring, Christian and Jud, Dominic and Lauber, Andreas and Bellicoso, C Dario and Tsounis, Vassilios and Hwangbo, Jemin and Bodie, Karen and Fankhauser, Peter and Bloesch, Michael and others},
  booktitle={2016 IEEE/RSJ international conference on intelligent robots and systems (IROS)},
  pages={38--44},
  year={2016},
  organization={IEEE}
}

@online{unitree,
  author    = {unitree},
  title     = {About the unitree Robot},
  year      = {2023},
  url       = {https://shop.unitree.com/products/unitree-go2?srsltid=AfmBOopSkw67HujLhIwAHpq1DLuCBe7h4Qh_z4c4EaotY6eFRrMvbPo8
}
}

@online{bostondynamics2023,
  author    = {Boston Dynamics},
  title     = {About the Spot Robot},
  year      = {2023},
  url       = {https://support.bostondynamics.com/s/article/About-the-Spot-Robot-72005},
  note      = {Accessed: 2024-09-15}
}

@article{miki2022learning,
  title={Learning robust perceptive locomotion for quadrupedal robots in the wild},
  author={Miki, Takahiro and Lee, Joonho and Hwangbo, Jemin and Wellhausen, Lorenz and Koltun, Vladlen and Hutter, Marco},
  journal={Science robotics},
  volume={7},
  number={62},
  pages={eabk2822},
  year={2022},
  publisher={American Association for the Advancement of Science}
}

@article{cheng2024navila,
  title={Navila: Legged robot vision-language-action model for navigation},
  author={Cheng, An-Chieh and Ji, Yandong and Yang, Zhaojing and Gongye, Zaitian and Zou, Xueyan and Kautz, Jan and B{\i}y{\i}k, Erdem and Yin, Hongxu and Liu, Sifei and Wang, Xiaolong},
  journal={arXiv preprint arXiv:2412.04453},
  year={2024}
}

@inproceedings{stolle2024perceptive,
  title={Perceptive pedipulation with local obstacle avoidance},
  author={Stolle, Jonas and Arm, Philip and Mittal, Mayank and Hutter, Marco},
  booktitle={2024 IEEE-RAS 23rd International Conference on Humanoid Robots (Humanoids)},
  pages={157--164},
  year={2024},
  organization={IEEE}
}

@article{sun2025learning,
  title={Learning Perceptive Humanoid Locomotion over Challenging Terrain},
  author={Sun, Wandong and Cao, Baoshi and Chen, Long and Su, Yongbo and Liu, Yang and Xie, Zongwu and Liu, Hong},
  journal={arXiv preprint arXiv:2503.00692},
  year={2025}
}

@article{blickhan1989spring,
  title={The spring-mass model for running and hopping},
  author={Blickhan, Reinhard},
  journal={Journal of biomechanics},
  volume={22},
  number={11-12},
  pages={1217--1227},
  year={1989},
  publisher={Elsevier}
}

@article{ijspeert2001connectionist,
  title={A connectionist central pattern generator for the aquatic and terrestrial gaits of a simulated salamander},
  author={Ijspeert, Auke Jan},
  journal={Biological cybernetics},
  volume={84},
  number={5},
  pages={331--348},
  year={2001},
  publisher={Springer}
}

@inproceedings{todorov2012mujoco,
  title={Mujoco: A physics engine for model-based control},
  author={Todorov, Emanuel and Erez, Tom and Tassa, Yuval},
  booktitle={2012 IEEE/RSJ international conference on intelligent robots and systems},
  pages={5026--5033},
  year={2012},
  organization={IEEE}
}

@inproceedings{pybullet,
  title={
Bullet Real-Time Physics Simulation},
  booktitle={https://pybullet.org/wordpress/},
}

@inproceedings{isaacsim,
  title={NVIDIA Isaac Sim},
  booktitle={https://developer.nvidia.com/isaac/sim},
  organization={NVIDIA}
}

@article{schulman2017proximal,
  title={Proximal policy optimization algorithms},
  author={Schulman, John and Wolski, Filip and Dhariwal, Prafulla and Radford, Alec and Klimov, Oleg},
  journal={arXiv preprint arXiv:1707.06347},
  year={2017}
}

@inproceedings{haarnoja2018soft,
  title={Soft actor-critic: Off-policy maximum entropy deep reinforcement learning with a stochastic actor},
  author={Haarnoja, Tuomas and Zhou, Aurick and Abbeel, Pieter and Levine, Sergey},
  booktitle={International conference on machine learning},
  pages={1861--1870},
  year={2018},
  organization={Pmlr}
}

@article{ewen2022these,
  title={These maps are made for walking: Real-time terrain property estimation for mobile robots},
  author={Ewen, Parker and Li, Adam and Chen, Yuxin and Hong, Steven and Vasudevan, Ram},
  journal={IEEE Robotics and Automation Letters},
  volume={7},
  number={3},
  pages={7083--7090},
  year={2022},
  publisher={IEEE}
}

@article{chen2024design,
  title={Design of an Adaptive Lightweight Lidar to decouple robot--camera geometry},
  author={Chen, Yuyang and Wang, Dingkang and Thomas, Lenworth and Dantu, Karthik and Koppal, Sanjeev J},
  journal={IEEE Transactions on Robotics},
  volume={40},
  pages={2254--2271},
  year={2024},
  publisher={IEEE}
}

@inproceedings{neuhaus2018mc2slam,
  title={Mc2slam: Real-time inertial lidar odometry using two-scan motion compensation},
  author={Neuhaus, Frank and Ko{\ss}, Tilman and Kohnen, Robert and Paulus, Dietrich},
  booktitle={German Conference on Pattern Recognition},
  pages={60--72},
  year={2018},
  organization={Springer}
}

@inproceedings{gojcic2019perfect,
  title={The perfect match: 3d point cloud matching with smoothed densities},
  author={Gojcic, Zan and Zhou, Caifa and Wegner, Jan D and Wieser, Andreas},
  booktitle={Proceedings of the IEEE/CVF conference on computer vision and pattern recognition},
  pages={5545--5554},
  year={2019}
}

@article{taketomi2017visual,
  title={Visual SLAM algorithms: A survey from 2010 to 2016},
  author={Taketomi, Takafumi and Uchiyama, Hideaki and Ikeda, Sei},
  journal={IPSJ transactions on computer vision and applications},
  volume={9},
  number={1},
  pages={16},
  year={2017},
  publisher={Springer}
}

@article{bazeille2017active,
  title={Active camera stabilization to enhance the vision of agile legged robots},
  author={Bazeille, St{\'e}phane and Ortiz, Jesus and Rovida, Francesco and Camurri, Marco and Meguenani, Anis and Caldwell, Darwin G and Semini, Claudio},
  journal={Robotica},
  volume={35},
  number={4},
  pages={942--960},
  year={2017},
  publisher={Cambridge University Press}
}

@article{wangy2025terrain,
  title={Terrain-Adaptive Planning of a Mobile Robot with a Multi-Axis Gimbal System for Stable SLAM},
  author={Wangy, Zhihao and Liy, Minghang and Liu, Xiao and Wang, Yu and Liu, Yiming and Chen, Haoyao},
  journal={IEEE Transactions on Field Robotics},
  year={2025},
  publisher={IEEE}
}

@article{chen2406slr,
  title={SLR: Learning Quadruped Locomotion without Privileged Information. arXiv 2024},
  author={Chen, S and Wan, Z and Yan, S and Zhang, C and Zhang, W and Liu, Q and Zhang, D and Farrukh, FD},
  journal={arXiv preprint arXiv:2406.04835}
}

@inproceedings{rudin2022learning,
  title={Learning to walk in minutes using massively parallel deep reinforcement learning},
  author={Rudin, Nikita and Hoeller, David and Reist, Philipp and Hutter, Marco},
  booktitle={Conference on robot learning},
  pages={91--100},
  year={2022},
  organization={PMLR}
}
\end{document}